\documentclass[pdflatex,sn-mathphys-num]{sn-jnl}

\usepackage{graphicx}
\usepackage{multirow}
\usepackage{amsmath,amssymb,amsfonts}
\usepackage{amsthm}
\usepackage{mathrsfs}
\usepackage[title]{appendix}
\usepackage{xcolor}
\usepackage{textcomp}
\usepackage{manyfoot}
\usepackage{booktabs}
\usepackage{algorithm}
\usepackage{algorithmicx}
\usepackage{algpseudocode}
\usepackage{listings}

\usepackage{graphicx}
\usepackage{colortbl}

\usepackage{amssymb}
\usepackage{amsmath}

\newcommand{\veryshortarrow}[1][3pt]{\mathrel{%
   \hbox{\rule[\dimexpr\fontdimen22\textfont2-.2pt\relax]{#1}{.4pt}}%
   \mkern-4mu\hbox{\usefont{U}{lasy}{m}{n}\symbol{41}}}}

\theoremstyle{thmstyleone}%
\theoremstyle{thmstyletwo}%

\theoremstyle{thmstylethree}%

\begin{document}

\title[Article Title]{Efficient Unsupervised Domain Adaptation via Self-Supervised Vision Transformer and Synergistic Cross-Domain Alignment}


\author*[1]{\fnm{Ali} \sur{Abedi}}\email{abedi3@uwindsor.ca}

\author[1]{\fnm{Q. M. Jonathan} \sur{Wu}}\email{jwu@uwindsor.ca}

\author[1]{\fnm{Ning} \sur{Zhang}}\email{ning.zhang@uwindsor.ca}

\author[2]{\fnm{Farhad} \sur{Pourpanah}}\email{farhad.086@gmail.com}

\affil*[1]{\orgdiv{Department of Electrical and Computer Engineering}, \orgname{University of Windsor}, \orgaddress{\city{Windsor}, \state{Ontario}, \country{Canada}}}

\affil[2]{\orgdiv{Department of Electrical and Computer Engineering}, \orgname{Queens
University}, \orgaddress{\city{Kingston}, \state{Ontario}, \country{Canada}}}


\abstract{
{\color{black}
Unsupervised domain adaptation (UDA) aims to mitigate domain shift, where the distribution of labeled source data differs from that of unlabeled target data. Despite recent advances, existing methods often rely on fine-tuning large backbone models, which leads to high computational cost and limits scalability in resource-constrained environments. This limitation highlights the need for parameter -efficient approaches that maintain strong performance with reduced training complexity. Self-supervised foundation models such as DINOv2 provide highly transferable representations and raise the question of whether effective domain adaptation can be achieved without full fine-tuning. To address this question, we propose \textit{\textbf{E}fficient \textbf{U}nsupervised \textbf{D}omain \textbf{A}daptation} (EUDA), a parameter-efficient framework that leverages a frozen DINOv2 backbone as a feature extractor and updates only a lightweight bottleneck and classification head. We also adopt a synergistic domain alignment loss (SDAL), which combines cross-entropy (CE) and maximum mean discrepancy (MMD) to promote both discriminative learning and cross-domain alignment. Experimental results on Office-Home, Office-31, VisDA-2017, and DomainNet demonstrate that EUDA achieves competitive performance across diverse domain complexities, while reducing the number of trainable parameters by 42\% to 99.7\%. These results show the suitability of the proposed method for resource-constrained and distributed environments.
}
}




\maketitle

\section{Introduction}\label{sec1}

Unsupervised domain adaptation (UDA) has shown promising results in addressing the domain shift problem, where the distribution of the training data (source domain) differs from that of the test data (target domain), and enabling knowledge transfer \cite{redko_advances_2019, kim_distilling_2022, ngo_improved_2023}. This discrepancy can significantly degrade the performance of conventional deep neural networks (DNNs) when applied directly to the target domain \cite{wang_deep_2018}. UDA overcomes this challenge by leveraging unlabeled data from the target domain to align feature distributions and improve generalization \cite{han_learning_2022}.
Additionally, traditional DNNs often rely on a large number of annotated samples, which limits their applicability in scenarios where labeling is expensive, such as autonomous driving \cite{li_domain_2023}. UDA mitigates this limitation by transferring knowledge from labeled source data to unlabeled target data, which reduces annotation costs while improving performance.

{\color{black}
Existing UDA methods can be broadly categorized into: (\textit{i}) adversarial methods, which learn domain-invariant features using generative adversarial training \cite{ganin_unsupervised_2015, tzeng_adversarial_2017, cao_partial_2018, ngo_higda_2025}; (\textit{ii}) reconstruction methods, which employ encoder–decoder architectures to learn invariant representations \cite{ghifary_domain_2015, bousmalis_domain_2016}; (\textit{iii}) transformation methods, which enhance adaptation through input transformations \cite{alkhalifah_direct_2021, han_learning_2022}; and (\textit{iv}) discrepancy-based methods, which align feature distributions using statistical measures such as maximum mean discrepancy (MMD) \cite{gretton_kernel_2006, long_learning_2015, yan_mind_2017, gilo_unsupervised_2023, gilo_subdomain_2024} and correlation alignment (CORAL) \cite{sun_return_2016, hua_deep_2016}. Across these approaches, the choice of feature extraction backbone plays a critical role, as more transferable and domain-invariant representations lead to improved adaptation performance.

Recently, vision transformers (ViTs) \cite{dosovitskiy_image_2021} have demonstrated strong performance in computer vision by modeling global dependencies through patch-based representations, which offers advantages over convolutional neural networks (CNNs) \cite{mauricio_comparing_2023}. This capability has made them particularly effective for UDA tasks \cite{yang_tvt_2021, xu_cdtrans_2022, zhu_patch-mix_2023, ngo_learning_2024}. Moreover, advances in self-supervised learning (SSL) have further enhanced the representation power of ViTs by enabling training on large-scale unlabeled data \cite{sariyildiz_concept_2021, han_survey_2023, oquab_dinov2_2024}. Among these, DINOv2 \cite{oquab_dinov2_2024} stands out for its ability to extract highly transferable and general-purpose features, which makes it a strong candidate for domain adaptation.

Despite these advances, most existing UDA methods rely on fully fine-tuning large backbone models, resulting in substantial computational cost and a large number of trainable parameters. This not only limits their applicability in resource-constrained environments but also makes them inefficient for distributed and large-scale scenarios such as federated learning, remote sensing, and medical image analysis, where communication cost, scalability, and computational constraints are critical. Therefore, there is a need for parameter-efficient UDA frameworks that leverage powerful pre-trained models while minimizing the number of trainable parameters.

In this paper, we introduce a novel \sloppy \textit{\textbf{E}fficient \textbf{U}nsupervised\textbf{ D}omain \textbf{A}daptation} framework, known as EUDA, a parameter-efficient framework that uses a frozen DINOv2 backbone as a feature extractor and adapts only a lightweight bottleneck composed of fully connected layers along with a classification head. Additionally, EUDA utilizes the synergistic domain alignment loss (SDAL), which combines cross-entropy (CE) loss and MMD loss. This hybrid loss minimizes classification error on the source domain and aligns source and target feature distributions, as illustrated in Figure~\ref{fig_MMD}. By restricting learning to a small subset of parameters, EUDA significantly reduces computational complexity while maintaining strong adaptation capability. Extensive experiments demonstrate that EUDA achieves competitive performance compared to state-of-the-art UDA methods while reducing the number of trainable parameters by 42\% to 99.7\%. This highlights the effectiveness of parameter-efficient adaptation and its potential for real-world applications where computational resources and communication costs are critical.

\begin{figure}[htbp]
\centering
\includegraphics[width=0.7\textwidth]{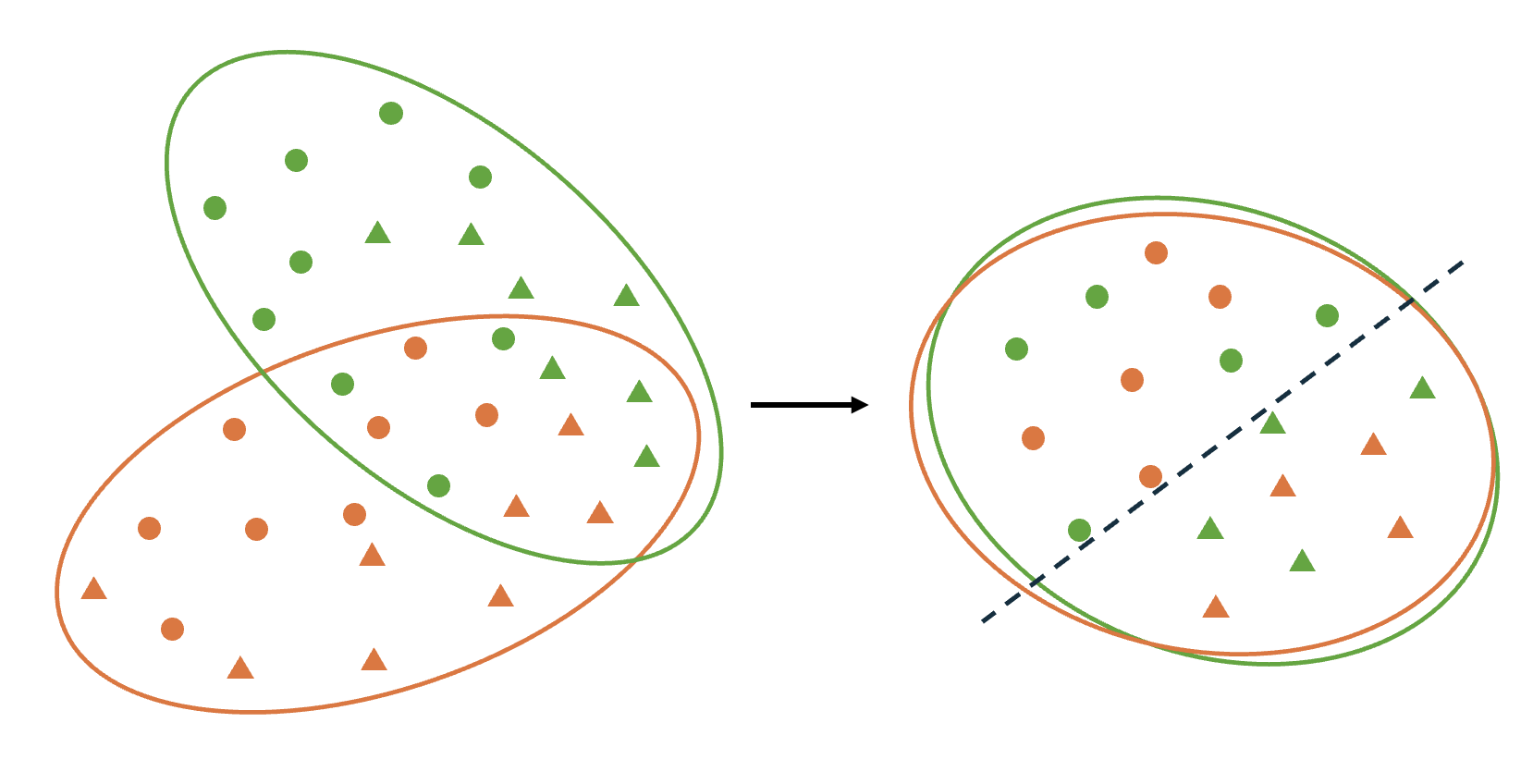}
\caption{{\color{black}The conceptual diagram illustrates how SDAL reduces the discrepancy between source and target data distributions while simultaneously improving classification accuracy. \textbf{Green} represents the source domain distribution, while \textbf{orange} represents the target domain distribution. \textbf{Circular} and \textbf{triangular} markers denote samples from two different classes. By combining MMD with CE loss, SDAL achieves both domain alignment and improved class discrimination.}}
\label{fig_MMD}
\end{figure}

In summary, the main contributions of this paper are as follows:
\begin{itemize}
\item We leverage the self-supervised vision transformer DINOv2 as a frozen feature extractor for unsupervised domain adaptation, which exploits its strong transferable representations to improve cross-domain generalization.

\item We propose EUDA (Efficient Unsupervised Domain Adaptation), a parameter-efficient framework that adapts a frozen foundation model using a lightweight bottleneck composed of fully connected layers. This design significantly reduces the number of trainable parameters while allowing the model complexity to be adjusted according to task requirements.

\item We introduce a synergistic domain alignment loss (SDAL) that integrates CE and MMD within a constrained adaptation framework, where both losses operate at the bottleneck level to preserve class-discriminative structure and align source and target representations without updating the backbone. This design enables effective domain alignment under limited trainable capacity.

\item We conduct extensive experiments and ablation studies across multiple benchmark datasets to demonstrate that EUDA achieves competitive performance while reducing trainable parameters by up to 99.7\% compared to fully fine-tuned transformer-based methods.
\end{itemize}

The rest of this paper is organized as follows: Section \ref{sec:related} reviews UDA, ViTs, and self-supervised learning approaches within ViTs. Section \ref{sec:methodology} details the proposed EUDA model. The experimental results across various configurations are presented in Section \ref{sec:experiment}. Finally, Section \ref{sec:conclusion} concludes the paper with insights and implications of our findings.
}

\section{Related Work}
\label{sec:related}
\subsection{Unsupervised Domain Adaptation}
As stated in the introduction, the domain shift problem is the major concern in UDA. Over the years, numerous methods have been developed to tackle this issue, aiming to enable effective model performance on the target domain without relying on labeled data \cite{tistarelli_domain_2023}. 

Adversarial methods use adversarial training to create domain-invariant features. For example, DANN \cite{ganin_unsupervised_2015} leverages a gradient reversal layer that inverts the gradient sign during training. ADDA \cite{tzeng_adversarial_2017} trains a source encoder on labeled images, and mixes source and target images to confuse the discriminator. Adversarial methods are computationally intensive, often exhibit unstable training processes, and do not always guarantee accurate feature mapping.

Transformation methods enhance domain adaptation by pre-processing input samples to optimize their condition for model training. DDA \cite{alkhalifah_direct_2021} preprocesses input data to align signal-to-noise ratios and reduce domain shifts, while TransPar \cite{han_learning_2022} applies the lottery ticket hypothesis to identify and adjust transferable network parameters for better cross-domain generalization. These approaches are simple and effective but may not capture all domain differences.

Reconstruction methods utilize an encoder-decoder setup to harmonize features across domains by reconstructing target domain images from source domain data. MTAE \cite{ghifary_domain_2015} utilized a multitask autoencoder to reconstruct images from multiple domains. DSNs \cite{bousmalis_domain_2016} enhance model performance by dividing image representations into domain-specific and shared subspaces. This improves generalization and surpasses other adaptation methods. They usually face challenges with high computational costs, training instability, and potential overfitting to the source domain.

Discrepancy methods have emerged as particularly effective for UDA. DAN \cite{long_learning_2015} embeds task-specific layer representations into a reproducing kernel Hilbert space (RKHS) and uses MMD to explicitly match the mean embeddings of different domain distributions. WeightedMMD \cite{yan_mind_2017} introduces a weighted MMD that incorporates class-specific auxiliary weights to address class weight bias in domain adaptation. This approach optimizes feature alignment between source and target domains by considering class prior distributions. Joint adaptation networks \cite{long_deep_2017} align the joint distributions of multiple domain-specific layers across domains using a Joint MMD criterion to improve domain adaptation by considering the combined shift in input features and output labels.

\subsection{Vision Transformer}
Transformers, initially introduced by Vaswani \textit{et al.} \cite{vaswani_attention_2017}, have demonstrated exceptional performance across various language tasks. The core of their success lies in the attention mechanism, which excels at capturing long-range dependencies. ViT \cite{dosovitskiy_image_2021} represents a groundbreaking approach to applying transformers in vision tasks. It treats 
images as sequences of fixed-size, non-overlapping patches. Unlike CNNs that depend on inductive biases such as locality and translation equivariance, ViT leverages the power of large-scale pre-training data and global context modeling. ViT offers a straightforward yet effective balance between accuracy and computational efficiency \cite{mauricio_comparing_2023}.

In the context of UDA, ViTs have demonstrated remarkable potential. TVT \cite{yang_tvt_2021} introduces a transferability adaptation module to guide the attention mechanism and a discriminative clustering technique to enhance feature diversity. CDTrans \cite{xu_cdtrans_2022} consists of a triple-branch structure with weight-sharing and cross-attention to align features from source and target domains, alongside a bidirectional center-guided pseudo labeling strategy to improve label quality. WinTR \cite{ma_exploiting_2021} uses two classification tokens within a transformer model to learn distinct domain mappings with domain-specific classifiers. This enhances the cross-domain knowledge transfer through source-guided label refinement and single-sided feature alignment. PMTrans \cite{zhu_patch-mix_2023} combines patches from both domains using game-theoretical principles, mixup losses, and attention maps for effective domain alignment and feature learning.

While these methods have shown promising performance in solving UDA problems, they typically rely on complex architectures with extensive trainable parameters and sophisticated training regimes, including multi-branch transformers, cross-attention, adversarial training, game-theoretical principles, and mixup losses. Furthermore, these models generally require training the entire network, resulting in a substantial computational burden. As a result, achieving promising outcomes necessitates extensive training on large-scale models, limiting their practical applicability in resource-constrained environments \cite{yang_tvt_2021, xu_cdtrans_2022, zhu_patch-mix_2023}.

\subsection{Self-supervised Vision Transformer}
SSL has revolutionized the field of computer vision by enabling models to learn effective representations from unlabeled data, eliminating the dependency on large annotated datasets. These models learn representations by performing pre-text tasks, such as rotation prediction \cite{feng_self-supervised_2019} and image colorization \cite{zhang_colorful_2016}, and then apply the learned representations to downstream tasks.
In the context of ViTs, SSL has been pivotal in enhancing their ability to extract robust, domain-invariant features. Jigsaw-ViT \cite{chen_jigsaw-vit_2023} integrates the jigsaw puzzle-solving problem into vision transformer architectures for improving image classification. EsViT \cite{li_efficient_2022} utilizes a multi-stage Transformer architecture to reduce computational complexity and introduces a novel region-matching pre-training task.

Among recent advancements in SSL for ViTs, DINO \cite{caron_emerging_2021} and DINOv2 \cite{oquab_dinov2_2024} have notably enhanced SSL by scaling up ViTs to effectively match representations across different views of the same image. With improvements like automatic data curation and innovative loss functions, DINOv2 excels in stability and efficiency. It can learn domain-invariant features essential for image classification and other vision tasks. Its capacity to generate robust feature maps makes DINOv2 an excellent choice for a feature extractor and representation generator. Because of these properties, we use DINOv2 in this research to significantly boost performance and generalization across various domains.

\section{Methodology}
\label{sec:methodology}
\subsection{Problem Formulation}
UDA aims to learn a function \( f: \mathcal{X} \to \mathcal{Y} \) that performs well on an unlabeled target domain by leveraging information from a labeled source domain. Let \( \mathcal{X} \) and \( \mathcal{Y} \) denote the input and label spaces, respectively, \( \mathcal{D}_s = \{(x_i^s, y_i^s)\}_{i=1}^{n_s} \) indicates the source domain data, where \( x_i^s \in \mathcal{X} \) and \( y_i^s \in \mathcal{Y} \) represent the $i^{th}$ input-output pairs, and \( \mathcal{D}_t = \{x_j^t\}_{j=1}^{n_t} \) indicates the target domain data, where \( x_j^t \in \mathcal{X} \) is the $j^{th}$ input samples without labels. Here, \( n_s \) and \( n_t \) represent the number of samples in the source and target domains, respectively. The goal of UDA is to train a model on labeled source data $\mathcal{D}_s$ and unlabeled target data $\mathcal{D}_t$ in such a way that it performs well on the predicting target data labels $y_i^t$.

\subsection{Model Overview}
\begin{figure}[tb]
\centering
\includegraphics[width=\textwidth]{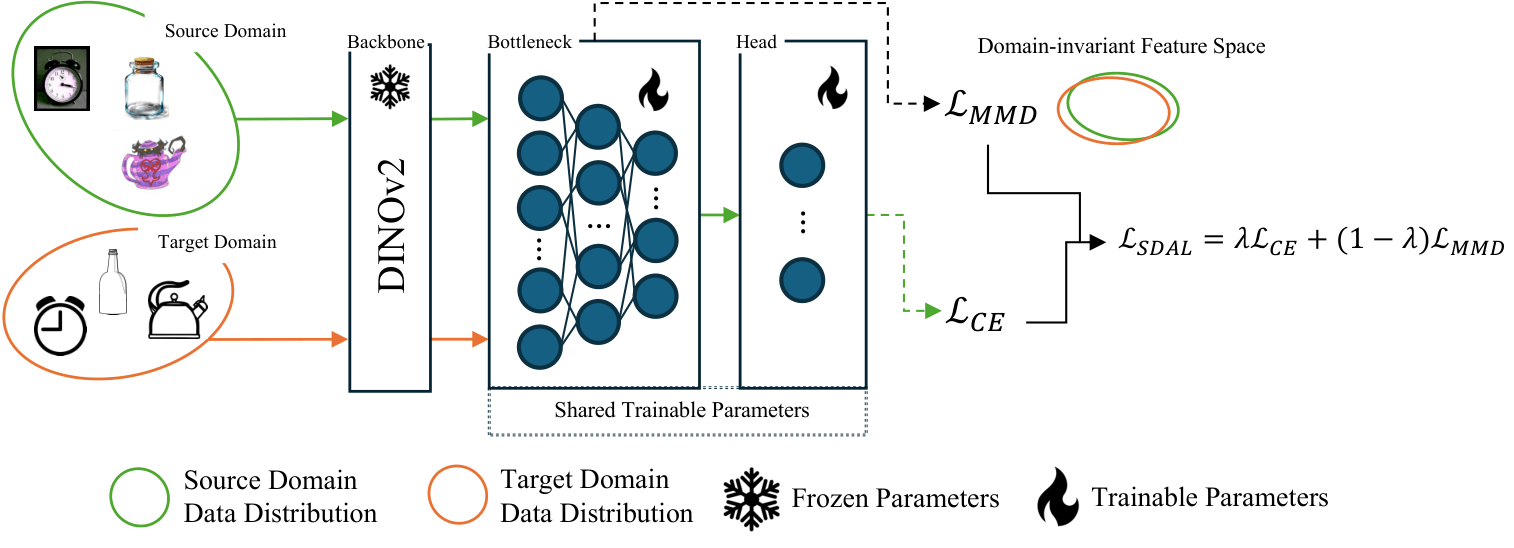}
\caption{{\color{black}The architecture of the proposed EUDA model. Labeled source and unlabeled target samples are processed by the frozen backbone $\phi$ to extract features, which are then refined by the bottleneck $\beta$. The refined features from both domains are used to compute the MMD loss for domain alignment, while source features are passed to the classification head $\psi$ to compute the cross-entropy loss.}}
\label{fig_MainModel}
\end{figure}

Figure~\ref{fig_MainModel} shows an overview of our framework for addressing UDA problems. 
The model receives images from labeled source and unlabeled target domains. {\color{black}A pre-trained self-supervised ViT, denoted as the feature extractor $\phi$, extracts features from both domains while its weights remain frozen to improve stability and computational efficiency. The extracted features are then passed to a bottleneck $\beta$, which consists of multiple fully connected layers and refines the representations into a compact feature space. The output of the bottleneck serves two purposes. First, it is forwarded to the classification head $\psi$ to compute the CE loss on the source domain.}
Second, it contributes to the computation of the MMD loss, which minimizes the distance between source and target domain distributions in RKHS and aligns their feature representations. Algorithm \ref{Algo_Training} presents the pseudocode of the EUDA training procedure. The following subsections describe each component of the model in detail.

\begin{algorithm}[ht]
\caption{The Training Procedure of EUDA}\label{alg:example}
\label{Algo_Training}
\begin{algorithmic}[1]
\State $\phi \leftarrow$ feature\_extractor with pre-trained DINOv2, freeze weights
\State $\beta \leftarrow$ bottleneck
\State $\psi \leftarrow$ classifier
\State $\theta \leftarrow$ optimizer
\State $\lambda \leftarrow$ lambda\_value
\State \textbf{for} epoch \textbf{in} range(num\_epochs) \textbf{do}
\State \ \ \ \ \textbf{for} $(x_s, y_s), x_t$ \textbf{in} DataLoader \textbf{do}
\State \ \ \ \ \ \ \ \ $f_s \leftarrow \phi(x_s)$
\State \ \ \ \ \ \ \ \ $f_t \leftarrow \phi(x_t)$
\State \ \ \ \ \ \ \ \ $\hat{y} \leftarrow \psi(\beta(f_s))$
\State \ \ \ \ \ \ \ \ $L_{ce} \leftarrow \text{cross\_entropy}(\hat{y}, y_s)$
\State \ \ \ \ \ \ \ \ $L_{mmd} \leftarrow \text{compute\_mmd}(\beta(f_s), \beta(f_t))$
\State \ \ \ \ \ \ \ \ $L \leftarrow \lambda \times L_{ce} + (1 - \lambda) \times L_{mmd}$
\State \ \ \ \ \ \ \ \ $\theta.\text{zero\_grad}()$
\State \ \ \ \ \ \ \ \ $L.\text{backward}()$
\State \ \ \ \ \ \ \ \ $\theta.\text{step}()$
\State \ \ \ \ \textbf{end for}
\State \textbf{end for}
\State Evaluate on target data
\end{algorithmic}
\end{algorithm}

\subsection{Feature Extractor}
In our effort to leverage the capabilities of ViTs for domain adaptation, we adopt DINOv2 \cite{oquab_dinov2_2024} as a self-supervised pre-trained model for feature extraction. It utilizes self-distillation to derive insights from unlabeled data autonomously. Central to DINOv2's design is its twin-network structure, which includes a student and a teacher network. Both networks employ the same underlying architecture based on ViTs. During training, these networks process different augmentations of the same image. They aim to extract consistent features regardless of the input variations.

\begin{figure}[htbp]
\centering
\includegraphics[width=0.4\textwidth]{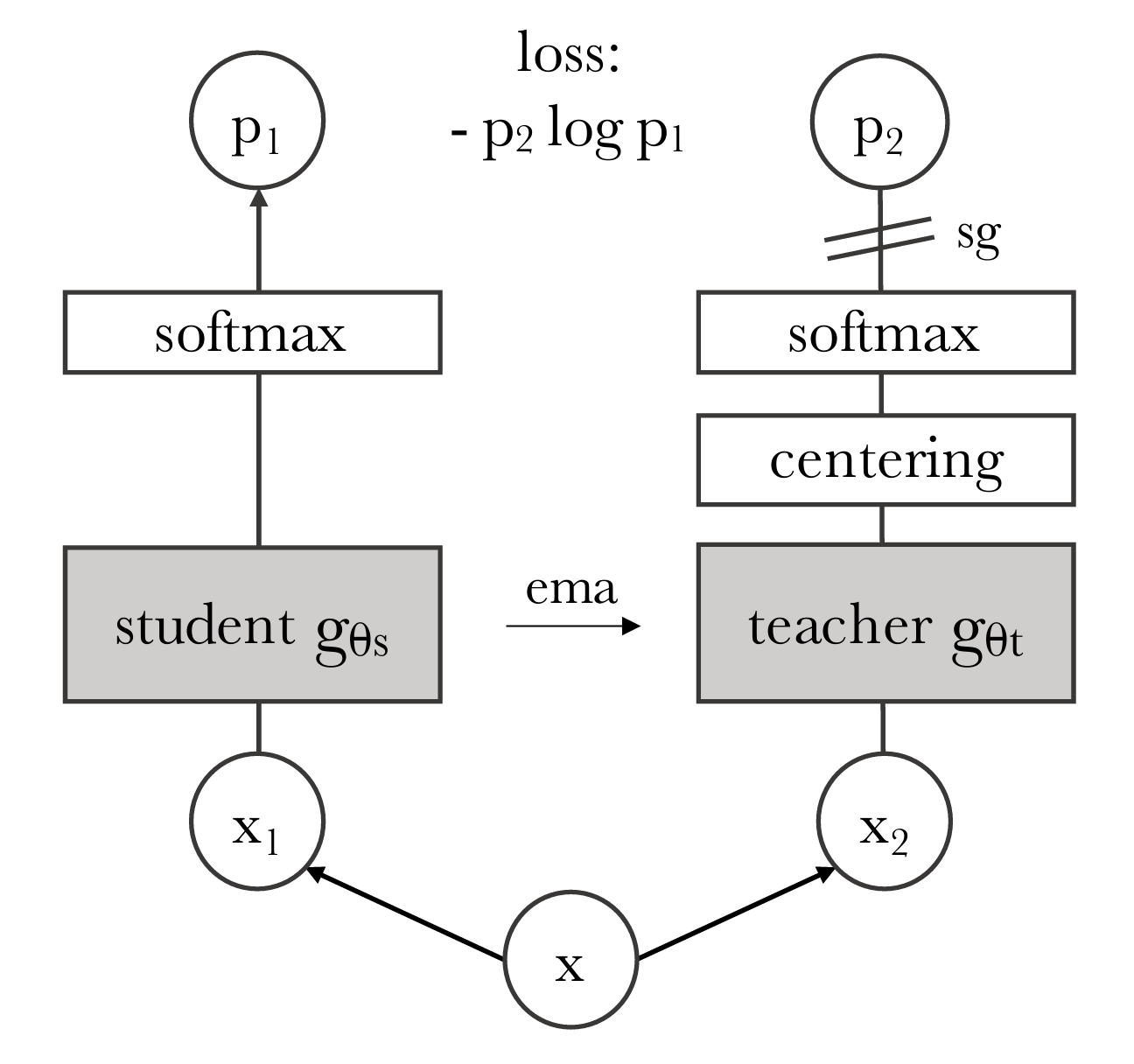}
\caption{Self-distillation with no labels. Image from \cite{caron_emerging_2021}.}
\label{fig_dinov2}
\end{figure}

During the training phase (see Figure~\ref{fig_dinov2}), the student network's parameters are continually updated, while the teacher network's parameters are progressively updated through an exponential moving average of the student's parameters. This ensures that the teacher model remains robust and generalizable. Moreover, DINOv2 uses registers \cite{darcet_vision_2024} to improve the performance and interpretability of ViTs by addressing the problem of artifacts in feature maps, commonly observed in both supervised and self-supervised ViT networks. Registers are additional tokens added to the input sequence of ViTs to absorb high-norm token computations that typically occur in low-information areas of images. 

\begin{figure}[htbp]
\centering
\includegraphics[width=0.6\textwidth]{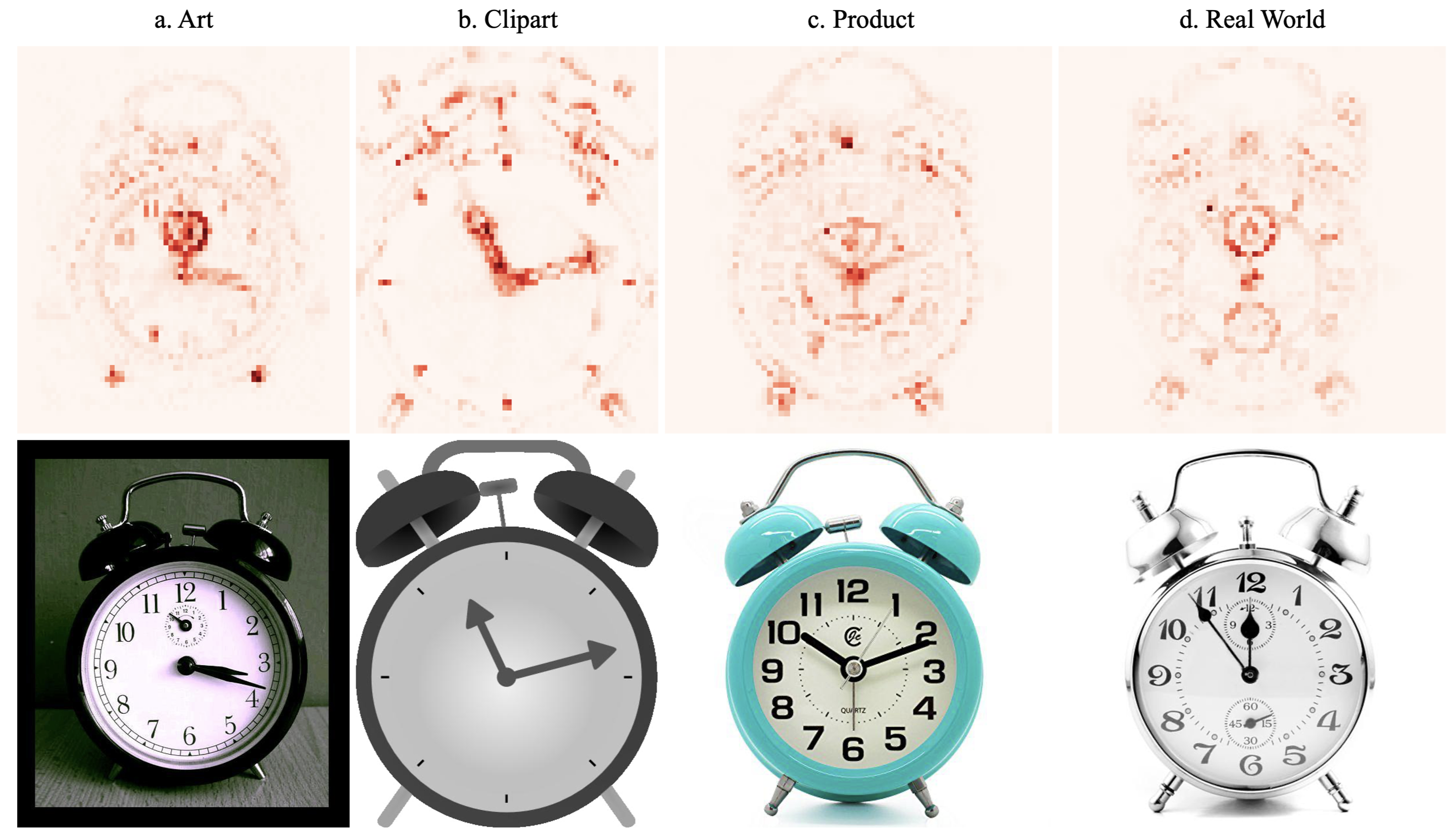}
\caption{Attention maps of the pre-trained DINOv2 base model on an Alarm Clock from four different domains from the office-home dataset: Art, Clipart, Product, and Real World. This illustrates the model's robust feature extraction capability across diverse image contexts without any fine-tuning.}
\label{fig_AM_OH}
\end{figure}

In our study, we leverage DINOv2 as the primary feature extractor due to its robust training on a large-scale, diverse dataset through self-supervised learning. To enhance efficiency, we freeze the model's parameters. This approach reduces the computational burden and significantly decreases the number of trainable parameters. This makes our method notably more efficient compared to other UDA techniques that require extensive training. Consequently, our streamlined model is well-suited for deployment in real-world scenarios and on-edge devices, where computational resources are often limited. Figure~\ref{fig_AM_OH} shows the attention map of the pre-trained DINOv2 base model without any fine-tuning. This highlights its robustness across four different domains of the office-home \cite{venkateswara_deep_2017} dataset. This demonstrates the precision and accuracy of the features extracted by DINOv2, underscoring the effectiveness of the EUDA feature extraction approach.

Using DINOv2 as the feature extractor significantly enhances our model by employing its robust, self-supervised pre-training to extract general-purpose features from images. The pre-trained DINOv2, with its weights frozen, ensures the extraction of high-quality features and reduces the number of trainable parameters. It can greatly simplify integration and adaptation in resource-limited settings, providing a strong foundation for effective domain adaptation.

\subsection{Bottleneck}
{\color{black}
As shown in Figure~\ref{fig_MainModel}, the bottleneck $\beta$ is positioned between the feature extractor $\phi$ and the classifier $\psi$. It processes features from both source and target domains and maps them into a representation suitable for classification and domain alignment.

The bottleneck consists of a sequence of fully connected layers, where the number of layers and hidden dimensions can be adjusted to control the model capacity. Its output is used for two purposes: (1) source features are passed to the classifier $\psi$ for supervised learning via CE loss, and (2) source and target features are used to compute the MMD loss for domain alignment.

This design provides a compact and flexible transformation module that refines features for both tasks while maintaining parameter efficiency.}

{\color{black}
\subsection{Synergistic Domain Alignment Loss}

In EUDA, the feature extractor $\phi$ is initialized with a pre-trained DINOv2 model and kept frozen during training. Therefore, domain adaptation is achieved through the trainable bottleneck $\beta$ and classifier $\psi$.

Given source samples $(x_s, y_s)$ and target samples $x_t$, features are first extracted using the frozen backbone:
\begin{equation}
f_s = \phi(x_s), \qquad f_t = \phi(x_t).
\end{equation}

These features are then passed through the bottleneck:
\begin{equation}
z_s = \beta(f_s), \qquad z_t = \beta(f_t).
\end{equation}

The classifier $\psi$ produces predictions on the source domain:
\begin{equation}
\hat{y} = \psi(z_s).
\end{equation}

To preserve discriminative performance on the labeled source domain, we use the CE loss:
\begin{equation}
\mathcal{L}_{CE} = -\frac{1}{N} \sum_{i=1}^{N} \sum_{c=1}^{C} y_{i,c} \log(\hat{y}_{i,c}),
\end{equation}
where $N$ is the number of source samples and $C$ is the number of classes.

To minimize the discrepancy between source and target domains, we apply the MMD loss on the bottleneck representations:
\begin{equation}
\mathcal{L}_{MMD} =
\left\|
\frac{1}{n_s}\sum_{i=1}^{n_s}\phi(z_i^s) -
\frac{1}{n_t}\sum_{j=1}^{n_t}\phi(z_j^t)
\right\|_{\mathcal{H}}^2,
\end{equation}
where $n_s$ and $n_t$ denote the number of source and target samples, respectively, and $\phi(\cdot)$ maps features into a reproducing kernel Hilbert space.

The final SDAL is defined as:
\begin{equation}
\mathcal{L}_{SDAL} = \lambda \mathcal{L}_{CE} + (1-\lambda)\mathcal{L}_{MMD},
\label{eq:loss}
\end{equation}
where $\lambda$ controls the balance between source supervision and domain alignment.

In EUDA, the feature extractor $\phi$ is frozen, and adaptation is performed solely through the lightweight bottleneck $\beta$ and classifier $\psi$. Unlike conventional UDA methods that rely on updating the entire backbone, EUDA enforces domain adaptation within a constrained, low-capacity feature space. In this setting, $\mathcal{L}_{CE}$ preserves class-discriminative structure on the source domain, while $\mathcal{L}_{MMD}$ aligns source and target representations at the bottleneck level. This design enables effective domain adaptation without backbone fine-tuning, making the approach both parameter-efficient and robust.
}

\section{Experimental Studies}
\label{sec:experiment}

\subsection{Datasets}
We examine the effectiveness of our proposed method across three benchmarks. These datasets include: 
\textbf{Office-31} \cite{hutchison_adapting_2010}, which consists of 4,652 images from three domains: Amazon, Webcam, and DSLR, across 31 categories; \textbf{Office-Home} \cite{venkateswara_deep_2017}, which consists of 15,500 images spread over four domains: Art, Clipart, Products, and Real World, within 65 categories; {\color{black}\textbf{VisDA-2017} \cite{peng_visda_2017}consists of a large-scale synthetic-to-real domain adaptation task with 12 categories, where synthetic images serve as the source domain and real images serve as the target domain;} and \textbf{DomainNet} \cite{peng_moment_2019}, which consists of 48,129 images from six domains: Clipart, Real, Sketch, Infograph, Painting, and Quickdraw, across 345 categories. 

\subsection{Evaluation}
To evaluate the effectiveness of our UDA model, following the procedure in \cite{yang_tvt_2021, xu_cdtrans_2022, zhu_patch-mix_2023}, We train our model by alternately using each domain as the source with labeled data and another as the target with unlabeled data, then evaluate using labeled data from the target domain. This procedure is repeated until all domains have been used as the source domain. This setup ensures that the model's ability to generalize to new, unseen environments is properly tested. The main evaluation criterion is accuracy, measured by the model’s ability to correctly classify unseen samples after being trained with source domain data and assessed on target domain data. This metric provides a clear measure of the model's performance in bridging the gap between disparate data distributions of the source and target domains.

\subsection{Baselines}
We compare our model against a broad spectrum of state-of-the-art models. We categorize these models into ResNet- and ViT-based models. 
The ResNet-based models include RevGrad \cite{ganin_unsupervised_2015}, CDAN \cite{long_conditional_2018}, TADA \cite{liu_transferable_2019}, SHOT \cite{liang_we_2020}, JAN \cite{long_deep_2017}, BNM \cite{cui_towards_2020}, MCD \cite{saito_maximum_2018}, SWD \cite{lee_sliced_2019}, and DTA \cite{lee_drop_2019}. While ViT-based models include CDTrans \cite{xu_cdtrans_2022}, PMTrans \cite{zhu_patch-mix_2023}, TVT \cite{yang_tvt_2021}, and PCaM \cite{zang_pcam_2025}. 

\subsection{Implementation Details}
In our domain adaptation model, we designed the bottleneck component with varying complexities to assess how architectural depth influences feature processing. Our configurations ranged from a simple single-layer network with 256 neurons (designated S for Small) to more complex setups like 2048-1024-512-256 (B for Base model), 4096-2048-1024-512-256 (L for Large model), and 8192-4096-2048-1024-512-256 (H for Huge model). We also employed different sizes of the DINOv2 model, base and large, as feature extractors to examine their effects on domain adaptation performance. We used a naming convention in XY format for clarity, where X represents the feature extractor size (B for base and L for large) and Y the bottleneck size. For instance, BB indicates that both the feature extractor and bottleneck are base size.

To optimize domain alignment and classification accuracy, we experimented with different values of the hyperparameter $\lambda$, and based on the experiments conducted on Section \ref{sec:lambdaSADL}, the value of 0.7 was selected for further experiments. {\color{black}Adjustments were made to the batch size based on model configuration and available GPU memory (1080Ti) to ensure stable training and fair comparison. For smaller models and datasets, a batch size of 32 was used to match baseline settings, while larger configurations required reduced batch sizes. The learning rate starts at 3e-2 and gradually decreases. The model's code is publicly available.}

\subsection{Comparison with the baseline models}
\textbf{Performance.} Tables \ref{table_comp_oh}, \ref{table_comp_o31}, \ref{table_comp_visda17} and \ref{table_comp_domainnet} show the accuracy rates of EUDA and baseline models on Office-Home, Office-31, VISDA-2017 and DomainNet, respectively. 
As can be seen, our model consistently outperformed all ResNet-based models across all datasets. Specifically, for the Office-Home dataset, our LL model surpassed CDTrans and TVT while delivering comparable results to PMTrans. For the Office-31 dataset, our model exceeded the performance of TVT and matched that of CDTrans and PMTrans. In the case of the VISDA-2017 dataset in Table \ref{table_comp_visda17}, our model performed closely to other SOTA models. For the DomainNet dataset, our findings were particularly striking. Despite the large number of classes the dataset contains, our smallest model configuration, the BS, outperformed most of the SOTA models, including CDTrans, and delivered results comparable to PMTrans. 

\begin{table*}[htbp]
\caption{Comparison of our best model with other methods on \textbf{Office-Home} dataset. The best performances and our model are marked as bold. \textbf{Note that our model achieved these results with about 83\% less learnable parameters than the best SOTA.}}
\label{table_comp_oh}
\centering
\def\arraystretch{1.5}
\resizebox{\textwidth}{!}{%
\begin{tabular}{l|l|lllllllllllll}
\toprule
Model          & Backbone & A $\veryshortarrow$ C &
  A $\veryshortarrow$ P &
  A $\veryshortarrow$ R &
  C $\veryshortarrow$ A &
  C $\veryshortarrow$ P &
  C $\veryshortarrow$ R &
  P $\veryshortarrow$ A &
  P $\veryshortarrow$ C &
  P $\veryshortarrow$ R &
  R $\veryshortarrow$ A &
  R $\veryshortarrow$ C &
  R $\veryshortarrow$ P &
  Avg.     \\ \midrule
Source Only    & ResNet       & 34.9  & 50.0  & 58.0  & 37.4  & 41.9  & 46.2  & 38.5  & 31.2  & 60.4  & 53.9  & 41.2          & 59.9  & 46.1  \\
RevGrad        & ResNet       & 45.6  & 59.3  & 70.1  & 47.0  & 58.5  & 60.9  & 46.1  & 43.7  & 68.5  & 63.2  & 51.8          & 76.8  & 57.6  \\
CDAN           & ResNet       & 50.7  & 70.6  & 76.0  & 57.6  & 70.0  & 70.0  & 57.4  & 50.9  & 77.3  & 70.9  & 56.7          & 81.6  & 65.8  \\
TADA           & ResNet       & 53.1  & 72.3  & 77.2  & 59.1  & 71.2  & 72.1  & 59.7  & 53.1  & 78.4  & 72.4  & 60.0          & 82.9  & 67.6  \\
SHOT           & ResNet       & 57.1  & 78.1  & 81.5  & 68.0  & 78.2  & 78.1  & 67.4  & 54.9  & 82.2  & 73.3  & 58.8          & 84.3  & 71.8  \\ \midrule
Source Only    & ViT (Deit)   & 55.6  & 73.0  & 79.4  & 70.6  & 72.9  & 76.3  & 67.5  & 51.0  & 81.0  & 74.5  & 53.2          & 82.7  & 69.8  \\
Source Only    & ViT          & 66.16 & 84.28 & 86.64 & 77.92 & 83.28 & 84.32 & 75.98 & 62.73 & 88.66 & 80.10 & 66.19         & 88.65 & 78.74 \\
TVT - Baseline & ViT          & 71.94 & 80.67 & 86.67 & 79.93 & 80.38 & 83.52 & 76.89 & 70.93 & 88.27 & 83.02 & 72.91         & 88.44 & 80.30 \\
CDTrans        & ViT (Deit)   & 68.8  & 85.0  & 86.9  & 81.5  & 87.1  & 87.3  & 79.6  & 63.3  & 88.2  & 82.0  & 66.0          & 90.6  & 80.5  \\
PCaM     & ViT        & 71.2 & 86.2 & 87.6 & 82.0 & 87.5 & 87.6 & 79.8 & 65.3 & 88.4 & 82.4 & 82.4 & 90.7 & 81.3 \\
TVT            & ViT          & 74.8 & 86.8 & 89.4 & 82.7 & 87.9 & 88.2 & 79.8 & 71.9 & 90.1 & 85.4 & 74.6         & 90.5 & 83.5 \\
\textbf{EUDA - LL (Ours)}      & ViT (DinoV2) & 80.6  & 84.9  & 88.4  & 85.2  & 88.0  & 88.6  & 76.6  & 77.4  & 86.7  & 87.7  & \textbf{82.5} & 92.8  & 84.9  \\
PMTrans &
  ViT &
  \textbf{81.2} &
  \textbf{91.6} &
  \textbf{92.4} &
  \textbf{88.9} &
  \textbf{91.6} &
  \textbf{93.0} &
  \textbf{88.5} &
  \textbf{80.0} &
  \textbf{93.4} &
  \textbf{89.5} &
  82.4 &
  \textbf{94.5} &
  \textbf{88.9}\\
  \bottomrule
\end{tabular}%
}
\end{table*}

\begin{table*}[htbp]
\caption{Comparison of our best model with other methods on \textbf{Office-31} dataset. The best performances and our model are marked as bold. \textbf{Note that our model achieved these results with about 83\% less learnable parameters than the best SOTA.}}
\label{table_comp_o31}
\centering
\def\arraystretch{1.5}
\resizebox{\textwidth}{!}{
\begin{tabular}{l|l|lllllll}
\toprule
Model          & Backbone & A $\veryshortarrow$ W & D $\veryshortarrow$ W & W $\veryshortarrow$ D & A $\veryshortarrow$ D & D $\veryshortarrow$ A & W $\veryshortarrow$ A & Avg.           \\ \midrule
Source Only    & ResNet       & 68.9          & 68.4           & 62.5           & 96.7          & 60.7          & 99.3           & 76.1          \\
RevGrad        & ResNet       & 82.0          & 96.9           & 99.1           & 79.7          & 68.2          & 67.4           & 82.2          \\
JAN            & ResNet       & 86.0          & 96.7           & 99.7           & 85.1          & 69.2          & 70.7           & 84.6          \\
CDAN           & ResNet       & 86.0          & 96.7           & 99.7           & 85.1          & 69.2          & 70.7           & 84.6          \\
SHOT           & ResNet       & 90.1          & 98.4           & 99.9           & 94.0          & 74.7          & 74.3           & 88.6          \\ \midrule
Source Only    & ViT (Deit)   & 90.8          & 90.4           & 76.8           & 98.2          & 76.4          & \textbf{100.0}           & 88.8          \\
Source Only    & ViT          & 89.2 & 98.9 & \textbf{100.0} & 88.8 & 80.1 & 79.8 & 89.5         \\
TVT - Baseline & ViT          & 91.6 & 99.0 & \textbf{100.0} & 90.6 & 80.2 & 80.2 & 90.3         \\
TVT            & ViT          & 91.6 & 99.0 & \textbf{100.0} & 90.6 & 80.2 & 80.2 & 90.3         \\
\textbf{EUDA - LL (Ours)}      & ViT (DinoV2) & 95.3         & \textbf{100.0} & \textbf{100.0} & 93.4          & 80.5          & 82.9           & 92.0          \\
CDTrans        & ViT (Deit)   & 97.0          & 96.7           & 81.1           & 99.0          & 81.9          & \textbf{100.0} & 92.6          \\
PMTrans        & ViT          & \textbf{99.1} & 99.6           & \textbf{100.0}          & \textbf{99.4} & \textbf{85.7} & 86.3           & \textbf{95.0}\\
\bottomrule
\end{tabular}
}
\end{table*}

\begin{table*}[htbp]
\caption{Comparison of our best model with other methods on \textbf{VisDA-2017} dataset. The best performance and our model are marked as bold. \textbf{Note that our model achieved these results with about 42\% less learnable parameters than the best SOTA.}}
\label{table_comp_visda17}
\centering
\def\arraystretch{1.5}
\resizebox{\textwidth}{!}{
\begin{tabular}{l|l|lllllllllllll}
\toprule
\textbf{Model}              & Backbone & plane & bcycl & bus   & car   & house & knife & mcycl & person & plant & sktbrd & train & truck & Avg.  \\ \midrule
Source Only        & ResNet     & 55.1  & 53.3  & 61.9  & 59.1  & 80.6  & 17.9  & 79.7  & 31.2   & 81.0  & 26.5   & 73.5  & 8.5   & 52.4  \\
RevGrad            & ResNet     & 81.9  & 77.7  & 82.8  & 44.3  & 81.2  & 29.5  & 65.1  & 28.6   & 51.9  & 54.6   & 82.8  & 7.8   & 57.4  \\
BNM                & ResNet     & 89.6  & 61.5  & 76.9  & 55.0  & 89.3  & 69.1  & 81.3  & 65.5   & 90.0  & 47.3   & 89.1  & 30.1  & 70.4  \\
MCD                & ResNet     & 87.0  & 60.9  & 83.7  & 64.0  & 88.9  & 79.6  & 84.7  & 76.9   & 88.6  & 40.3   & 83.0  & 25.8  & 71.9  \\
SWD                & ResNet     & 90.8  & 82.5  & 81.7  & 70.5  & 91.7  & 69.5  & 86.3  & 77.5   & 87.4  & 63.6   & 85.6  & 29.2  & 76.4  \\
DTA                & ResNet     & 93.7  & 82.2  & 85.6  & 83.8  & 93.0  & 81.0  & 90.7  & 82.1   & 95.1  & 78.1   & 86.4  & 32.1  & 81.5  \\
SHOT               & ResNet     & 94.3  & 88.5  & 80.1  & 57.3  & 93.1  & 94.9  & 80.7  & 80.3   & 91.5  & 89.1   & 86.3  & 58.2  & 82.9  \\ \midrule
Source Only & ViT (Deit) & 97.7  & 48.1  & 86.6  & 61.6  & 78.1  & 63.4  & 94.7  & 10.3   & 87.7  & 47.7   & 94.4  & 35.5  & 67.1  \\
Source Only        & ViT & 98.2 & 73.0 & 82.6 & 62.0 & 97.4 & 63.6 & 96.5 & 29.8 & 68.8 & 86.8 & 96.8 & 23.7 & 73.3 \\
TVT - Baseline     & ViT        & 94.6 & 81.6 & 81.9 & 69.9 & 93.6 & 70.0 & 88.6 & 50.5 & 86.8 & 88.5 & 91.5 & 20.1 & 76.5 \\
\textbf{EUDA - BH (Ours)} &
  ViT (DinoV2) & \textbf{99.5} & 78.1 & \textbf{90.6} & 58.1 & \textbf{98.5} & \textbf{98.5} & \textbf{97.8} & 63.4 & 79.8 & \textbf{97.3} & \textbf{98.2} & 37.1 & 83.2 \\
TVT & ViT & 93.0 & 85.6 & 77.6 & 60.5 & 93.6 & 98.2 & 89.4 & 76.4 & 93.6 & 92.1 & 91.7 & 55.8 & 84.0 \\
PMTrans      & ViT        & 98.2  & 92.2  & 88.1  & 77.0  & 97.4  & 95.8  & 94.0  & 72.1   & 97.1  & 95.2   & 94.6  & 51.0  & 87.7  \\
CDTrans &
  ViT (Deit) &
  97.1 &
  \textbf{90.5} &
  82.4 &
  \textbf{77.5} &
  96.6 &
  96.1 &
  93.6 &
  \textbf{88.6} &
  \textbf{97.9} &
  86.9 &
  90.3 &
  \textbf{62.8} &
  \textbf{88.4}\\
PCaM      & ViT        & 98.1  & 93.2  & 90.1  & 89.4  & 98.8  & 97.3  & 96.0  & 84.6   & 98.0  & 92.1   & 95.4  & 63.8  & 91.4  \\
  \bottomrule
\end{tabular}
}
\end{table*}

\begin{table*}[htbp]
\caption{Comparison of our best model with other methods on \textbf{DomainNet} dataset. The best performances and our model are marked as bold. \textbf{Note that our model achieved these results with about 99.7\% less learnable parameters than the best SOTA.}}
\label{table_comp_domainnet}
\centering
\def\arraystretch{1.5}
\resizebox{\textwidth}{!}{%
\begin{tabular}{c|lllllll|c|lllllll|c|lllllll}
\toprule
MCD &
  clp &   inf &   pnt &   qdr &   rel &   skt &  Avg. &   SWD &   clp &  inf &  pnt &  qdr &  rel &  skt &  Avg. &  BNM &  clp &   inf &   pnt &  qdr &  rel &  skt &  Avg. \\ \midrule
 clp &   - &   15.4 &   25.5 & 3.3 & 44.6 & 31.2 & 24.0 & clp & - & 14.7 & 31.9 & 10.1 & 45.3 & 36.5 & 27.7 & clp & - & 12.1 & 33.1 & 6.2 & 50.8 & 40.2 & 28.5 \\
inf & 24.1 & - & 24.0 & 1.6 & 35.2 & 19.7 & 20.9 & inf & 22.9 & - & 24.2 & 2.5 & 33.2 & 21.3 & 20.0 & inf & 26.6 & - & 28.5 & 2.4 & 38.5 & 18.1 & 22.8 \\
pnt &31.1 & 14.8 & - & 1.7 & 48.1 & 22.8 & 23.7 & pnt & 33.6 & 15.3 & - & 4.4 & 46.1 & 30.7 & 26.0 & pnt & 39.9 & 12.2 & - & 3.4 & 54.5 & 36.2 & 29.2 \\
qdr & 8.5 & 2.1 & 4.6 & - & 7.9 & 7.1 & 6.0 & qdr & 15.5 & 2.2 & 6.4 & - & 11.1 & 10.2 & 9.1 & qdr & 17.8 & 1.0 & 3.6 & - & 9.2 & 8.3 & 8.0 \\
rel & 39.4 & 17.8 & 41.2 & 1.5 & - & 25.2 & 25.0 & rel & 41.2 & 18.1 & 44.2 & 4.6 & - & 31.6 & 27.9 & rel & 48.6 & 13.2 & 49.7 & 3.6 & - & 33.9 & 29.8 \\
skt & 37.3 & 12.6 & 27.2 & 4.1 & 34.5 & - & 23.1 & skt & 44.2 & 15.2 & 37.3 & 10.3 & 44.7 & - & 30.3 & skt & 54.9 & 12.8 & 42.3 & 5.4 & 51.3 & - & 33.3 \\
Avg. & 28.1 & 12.5 & 24.5 & 2.4 & 34.1 & 21.2 & \cellcolor[HTML]{C0C0C0}20.5 & Avg. & 31.5 & 13.1 & 28.8 & 6.4 & 36.1 & 26.1 & \cellcolor[HTML]{C0C0C0}23.6 & Avg. & 37.6 & 10.3 & 31.4 & 4.2 & 40.9 & 27.3 & \cellcolor[HTML]{C0C0C0}25.3 \\ \midrule
CGDM & clp & inf & pnt & qdr & rel & skt & Avg. & MDD & clp & inf & pnt & qdr & rel & skt & Avg. & SCDA & clp & inf & pnt & qdr & rel & skt & Avg. \\ \midrule
clp & - & 16.9 & 35.3 & 10.8 & 53.5 & 36.9 & 30.7 & clp & - & 20.5 & 40.7 & 6.2 & 52.5 & 42.1 & 32.4 & clp & - & 18.6 & 39.3 & 5.1 & 55.0 & 44.1 & 32.4 \\
inf & 27.8 & - & 28.2 & 4.4 & 48.2 & 22.5 & 26.2 & inf & 33.0 & - & 33.8 & 2.6 & 46.2 & 24.5 & 28.0 & inf & 29.6 & - & 34.0 & 1.4 & 46.3 & 25.4 & 27.3 \\
pnt & 37.7 & 14.5 & - & 4.6 & 59.4 & 33.5 & 30.0 & pnt & 43.7 & 20.4 & - & 2.8 & 51.2 & 41.7 & 32.0 & pnt & 44.1 & 19.0 & - & 2.6 & 56.2 & 42.0 & 32.8 \\
qdr & 14.9 & 1.5 & 6.2 & - & 10.9 & 10.2 & 8.7 & qdr & 18.4 & 3.0 & 8.1 & - & 12.9 & 11.8 & 10.8 & qdr & 30.0 & 4.9 & 15.0 & - & 25.4 & 19.8 & 19.0 \\
rel & 49.4 & 20.8 & 47.2 & 4.8 & - & 38.2 & 32.0 & rel & 52.8 & 21.6 & 47.8 & 4.2 & - & 41.2 & 33.5 & rel & 54.0 & 22.5 & 51.9 & 2.3 & - & 42.5 & 34.6 \\
skt & 50.1 & 16.5 & 43.7 & 11.1 & 55.6 & - & 35.4 & skt & 54.3 & 17.5 & 43.1 & 5.7 & 54.2 & - & 35.0 & skt & 55.6 & 18.5 & 44.7 & 6.4 & 53.2 & - & 35.7 \\
Avg. & 36.0 & 14.0 & 32.1 & 7.1 & 45.5 & 28.3 & \cellcolor[HTML]{C0C0C0}27.2 & Avg. & 40.4 & 16.6 & 34.7 & 4.3 & 43.4 & 32.3 & \cellcolor[HTML]{C0C0C0}28.6 & Avg. & 42.6 & 16.7 & 37.0 & 3.6 & 47.2 & 34.8 & \cellcolor[HTML]{C0C0C0}30.3 \\ \midrule
 CDTrans & clp & inf & pnt & qdr & rel & skt & Avg. & \textbf{EUDA - BS (Ours)} & clp & inf & pnt & qdr & rel & skt & Avg. & PMTrans & clp & inf & pnt & qdr & rel & skt & Avg. \\ \midrule
clp & - & 29.4 & 57.2 & 26.0 & 72.6 & 58.1 & 48.7 & clp & - & 35.3 & 59.9 & 19.9 & 72.8 & 63.5 & 50.3 & clp & - & 34.2 & 62.7 & 32.5 & 79.3 & 63.7 & 54.5 \\
inf & 57.0 & - & 54.4 & 12.8 & 69.5 & 48.4 & 48.4 & inf & 64.9 & - & 57.3 &	15.0 & 72.4 & 56.9 & 53.3 & inf & 67.4 & - & 61.1 & 22.2 & 78.0 & 57.6 & 57.3 \\
pnt & 62.9 & 27.4 & - & 15.8 & 72.1 & 53.9 & 46.4 & pnt & 66.3 & 34.0 & - & 15.5 & 73.0 & 60.8 & 49.9 & pnt & 69.7 & 33.5 & - & 23.9 & 79.8 & 61.2 & 53.6 \\
qdr & 44.6 & 8.9 & 29.0 & - & 42.6 & 28.5 & 30.7 & qdr & 47.8 & 17.1 & 31.4 & - & 41.7 & 38.0 & 35.2 & qdr & 54.6 & 17.4 & 38.9 & - & 49.5 & 41.0 & 40.3 \\
rel & 66.2 & 31.0 & 61.5 & 16.2 & - & 52.9 & 45.6 & rel & 71.0 & 37.8 & 65.4 & 17.2 & - & 61.8 & 50.6 & rel & 74.1 & 35.3 & 70.0 & 25.4 & - & 61.1 & 53.2 \\
skt & 69.0 & 29.6 & 59.0 & 27.2 & 72.5 & - & 51.5 & skt & 72.5 & 35.7 & 62.3 & 19.2 & 72.4 & - & 52.4 & skt & 73.8 & 33.0 & 62.6 & 30.9 & 77.5 & - & 55.6 \\
Avg. & 59.9 & 25.3 & 52.2 & 19.6 & 65.9 & 48.4 & \cellcolor[HTML]{C0C0C0}45.2 & Avg. & 64.5 & 32.0 & 55.3 & 17.4 & 66.5 & 56.2 & \cellcolor[HTML]{C0C0C0} 48.7 & Avg. & 67.9 & 30.7 & 59.1 & 27.0 & 72.8 & 56.9 & \cellcolor[HTML]{C0C0C0}\textbf{52.4} \\ \bottomrule
\end{tabular}%
}
\end{table*}

{\color{black}
\textbf{Learnable parameters.} Table~\ref{table_lp} compares the trainable, frozen, and total parameters of EUDA with baseline transformer-based methods. Existing approaches such as TVT, PMTrans, and CDTrans fine-tune the entire backbone, resulting in over 85M trainable parameters, whereas EUDA freezes the pre-trained DINOv2 backbone and optimizes only a lightweight bottleneck and classification head. This design reduces the number of trainable parameters by 83\% on Office-Home and Office-31, 43\% on VisDA-2017, and up to 99.7\% on DomainNet, while maintaining competitive performance. Although the total parameter count remains large due to the frozen backbone, adaptation complexity is governed by trainable parameters, which determine optimization cost, memory usage, and communication overhead. By restricting learning to a small subset of parameters, EUDA enables efficient adaptation and is particularly well-suited for resource-constrained and distributed settings, such as federated learning, where minimizing sharable parameters significantly reduces communication cost and improves scalability.
}

\begin{table}[htbp]
\caption{Comparison of trainable, frozen, and total parameters for domain adaptation models.}
\label{table_lp}
\centering
\begin{tabular}{l|ccc}
\toprule
\textbf{Model} & \# Trainable Parameters & \# Frozen Parameters & \# Total Parameters \\ \hline
TVT (Base)     & 85.8M & 0 & 85.8M \\
PMTrans (ViT)  & 87.8M & 0 & 87.8M \\
PMTrans (Swin) & 90.0M & 0 & 90.0M \\
PMTrans (Deit) & 87.3M & 0 & 87.3M \\
CDTrans (Base) & 85.8M & 0 & 85.8M \\ \midrule
EUDA - BS      & 0.4M & 86M & 86.4M \\
EUDA - BB      & 4.4M & 86M & 90.4M   \\
EUDA - BL      & 14.3M & 86M & 100.3M  \\
EUDA - BH      & 51.1M & 86M & 137.1M  \\
EUDA - LS      & 0.4M & 300M & 300.4M   \\
EUDA - LB      & 4.4M & 300M & 304.4M   \\
EUDA - LL      & 14.3M & 300M & 314.3M  \\
EUDA - LH      & 51.1M & 300M & 351.1M  \\
\bottomrule
\end{tabular}
\end{table}

{\color{black}
\subsection{Ablation Study and Component Analysis }

In this section, we conduct an ablation study and component analysis to examine the contribution of the proposed components in EUDA.

\subsubsection{Ablation Study}
We removed the MMD component from Eq.~(\ref{eq:loss}) and training the model using only CE loss on the labeled source domain, referred to as \textit{Source Only}. In this setting, the model relies solely on source supervision without explicit target domain alignment.
As shown in Table~\ref{table_ours_oh}, on the Office-Home dataset, the BS and BL configurations achieve average accuracies of 79.8\% and 78.6\% under the source-only setting. Incorporating SDAL improves the performance to 81.0\% and 82.1\%, corresponding to gains of 1.2\% and 3.5\%, respectively. These results demonstrate that aligning feature distributions through MMD enhances generalization on the target domain.
A similar trend can be observed on the Office-31 and VisDA-2017 datasets (Tables~\ref{table_ours_o31} and \ref{table_ours_visda}), where incorporating SDAL consistently improves performance across different configurations.}

{\color{black}
\subsubsection{Impact of bottleneck configuration}
We further analyze the effect of the bottleneck architecture used in EUDA. The proposed framework employs a lightweight bottleneck composed of fully connected layers with varying capacities, denoted as Small (S), Base (B), Large (L), and Huge (H). Increasing the bottleneck capacity enables the effective refinement of frozen backbone features and facilitates the learning of stronger, domain-invariant representations.

As shown in Table~\ref{table_ours_oh}, increasing the bottleneck capacity generally improves performance. For example, the BS configuration achieves an average accuracy of 81.0\%, while the BB and BL configurations improve the performance to 82.2\% and 82.1\%, respectively. However, further increasing the bottleneck size to BH does not yield consistent gains. This observation suggests that a moderate bottleneck capacity is sufficient to capture the domain discrepancy while maintaining parameter efficiency.

\subsubsection{Impact of backbone scale}
We also investigate the influence of the backbone scale by comparing models based on the Base and Large variants of the DINOv2 backbone. As shown in Table~\ref{table_ours_oh}, models built on the Large backbone consistently achieve higher performance. For instance, the LL and LH configurations achieve the highest average accuracy of 84.9\% on Office-Home.

This improvement indicates that larger self-supervised representations provide more transferable features for domain adaptation. Importantly, since the backbone remains frozen in EUDA, increasing the backbone size does not increase the number of trainable parameters. Therefore, performance improvements can be achieved without increasing the adaptation complexity.

}

\begin{table*}[htbp]
\caption{Results on the \textbf{Office-Home} dataset showing the impact of SDAL (Source Only vs. full model), different bottleneck configurations (S, B, L, H), and backbone scales (Base and Large).}
\label{table_ours_oh}
\centering
\def\arraystretch{1.5}
\resizebox{\textwidth}{!}{%
\begin{tabular}{l|lllllllllllll}
\toprule
\textbf{Model}&
  A $\veryshortarrow$ C &
  A $\veryshortarrow$ P &
  A $\veryshortarrow$ R &
  C $\veryshortarrow$ A &
  C $\veryshortarrow$ P &
  C $\veryshortarrow$ R &
  P $\veryshortarrow$ A &
  P $\veryshortarrow$ C &
  P $\veryshortarrow$ R &
  R $\veryshortarrow$ A &
  R $\veryshortarrow$ C &
  R $\veryshortarrow$ P &
  Avg. \\ \midrule
BS - Source Only & 70.2 & 83.5 & 86.9 & 79.8 & 85.0 & 85.8 & 73.7 & 64.6 & 85.5 & 81.4 & 70.6 & 90.7 & 79.8 \\
BS & 73.9 & 84.0 & 86.9 & 79.4 & 85.1 & 85.5 & 74.5 & 68.6 & 85.5 & 82.2 & 75.9 & 90.5 & 81.0 \\
BB & 75.1 & 84.9 & 87.0 & 81.4 & 85.5 & 86.2 & 75.8 & 70.1 & 86.7 & 84.8 & 77.6 & 90.9 & 82.2 \\
BL - Source Only & 69.6 & 81.6 & 85.9 & 78.7 & 84.9 & 85.0 & 71.0 & 61.0 & 83.6 & 82.3 & 70.2 & 89.9 & 78.6 \\
BL & 75.5 & 84.0 & 87.3 & 80.7 & 85.1 & 86.0 & 78.5 & 69.9 & 86.4 & 84.4 & 77.1 & 90.8 & 82.1 \\
BH & 74.4 & 84.4 & 86.3 & 80.2 & 84.0 & 85.7 & 75.6 & 69.9 & 86.3 & 83.9 & 77.4 & 90.5 & 81.6 \\
LL & \textbf{80.6} & 84.9 & \textbf{88.4} & \textbf{85.2} & \textbf{88.0} & \textbf{88.6} & 76.6 & \textbf{77.4} & \textbf{86.7} & \textbf{87.7} & 82.5 & \textbf{92.8} & \textbf{84.9} \\
LH & 79.8 & \textbf{87.4} & 87.7 & 85.1 & 87.9 & 88.2 & \textbf{79.4} & 75.4 & 86.0 & 87.6 & \textbf{82.7} & 92.5 & \textbf{84.9}\\
\bottomrule
\end{tabular}
}
\end{table*}

\begin{table}[ht]
\caption{Results on the \textbf{Office-31} dataset showing the impact of SDAL (Source Only vs. full model), different bottleneck configurations (B, L), and backbone scales (Base and Large).}
\label{table_ours_o31}
\centering
\begin{tabular}{l|lllllll}
\toprule
\textbf{Model} & A $\veryshortarrow$ W & D $\veryshortarrow$ W & W $\veryshortarrow$ D & A $\veryshortarrow$ D & D $\veryshortarrow$ A & W $\veryshortarrow$ A & Avg.           \\ \midrule
BB - Source Only & 93.0 & 99.5 & 99.8 & 94.4 & 81.8 & 80.5 & 91.5 \\
BB & 94.9 & 99.4 & \textbf{100.0}& \textbf{95.2} & \textbf{80.6} & 81.1 & 91.9 \\
BL - Source Only & 93.6 & 99.2 & 99.8 & 93.0 & 80.3 & 79.4 & 90.9 \\
BL & 94.6 & 99.4 & \textbf{100.0}& 93.6 & 80.2 & 81.0 & 91.5 \\
LL & \textbf{95.3} & \textbf{100.0}& \textbf{100.0}& 93.4 & 80.5 & \textbf{82.9} & \textbf{92.0}\\
\bottomrule
\end{tabular}
\end{table}

\begin{table*}[htbp]
\caption{Results on the \textbf{Office-31} dataset showing the impact of SDAL (Source Only vs. full model), different bottleneck configurations (S, B, L, H), and backbone scales (Base and Large).}
\label{table_ours_visda}
\centering
\def\arraystretch{1.5}
\resizebox{\textwidth}{!}{%
\begin{tabular}{l|lllllllllllll}
\toprule
\textbf{Model}   & plane & bcycl & bus   & car   & house & knife & mcycl & person & plant & sktbrd & train & truck & Avg.  \\ \midrule
BB - Source Only & 97.9 & 79.6 & 91.3 & 56.1 & 88.5 & 65.9 & 96.2 & 22.0 & 74.4 & 92.9 & 94.3 & 33.7 & 74.4 \\
BB               & 99.4 & \textbf{78.1} & 90.7 & 56.9 & 98.5 & 97.7 & 97.6 & 61.3 & 77.2 & 97.2 & 97.7 & 37.2 & 82.5 \\
BL - Source Only & 98.9 & 77.9 & 90.8 & \textbf{59.2} & 93.9 & 63.7 & 94.8 & 37.0 & 74.1 & 86.8 & 91.7 & 33.5 & 75.2 \\
BL               & 99.5 & 77.5 & 91.0 & 55.9 & 98.3 & 98.0 & 97.5 & 62.5 & 78.4 & 97.3 & 98.0 & 36.9 & 82.6 \\
BH & 99.5 & \textbf{78.1} & 90.6 & 58.1 & \textbf{98.5} & 98.5 & \textbf{97.8} & \textbf{63.4} & \textbf{79.8} & 97.3 & 98.2 & 37.1 & \textbf{83.2} \\
LL & \textbf{99.9} & 72.6 & \textbf{91.1} & 55.6 & 97.8 & \textbf{98.8} & 97.7 & 50.8 & 61.9 & \textbf{98.6} & \textbf{98.6} & \textbf{39.4} & 80.2 \\
  \bottomrule
\end{tabular}
}
\end{table*}

\subsection{Sensitivity Analysis}
\subsubsection{Effect of $\lambda$}
\label{sec:lambdaSADL}
Table \ref{table_lambda} shows the impact of $\lambda$ in Eq. (\ref{eq:loss}) on the BS configuration for the Office-Home dataset.  In this experiment, we used four different values for $\lambda$: 0.3, 0.5, and 0.7. As can be seen, $\lambda = 0.7$ managed to produce the best results. The effectiveness of $\lambda = 0.7$ comes from its balanced approach that minimizes domain discrepancies through MMD loss while maintaining classification accuracy, ensuring robust performance across various domain shifts.

\begin{table*}[htb]
\caption{This table showcases the performance of our BS model with different $\lambda$ values (0.5, 0.3, and 0.7) on the Office-Home dataset across various domain combinations. The results confirm that $\lambda = 0.7$ consistently yields the highest classification accuracy, underscoring its effectiveness in balancing CE and MMD loss for optimal domain adaptation.}
\label{table_lambda}
\centering
\def\arraystretch{1.5}
\resizebox{\textwidth}{!}{%
\begin{tabular}{l|lllllllllllll}
\toprule
$\lambda$ &
  A $\veryshortarrow$ C &
  A $\veryshortarrow$ P &
  A $\veryshortarrow$ R &
  C $\veryshortarrow$ A &
  C $\veryshortarrow$ P &
  C $\veryshortarrow$ R &
  P $\veryshortarrow$ A &
  P $\veryshortarrow$ C &
  P $\veryshortarrow$ R &
  R $\veryshortarrow$ A &
  R $\veryshortarrow$ C &
  R $\veryshortarrow$ P &
  Avg. \\ \midrule
0.3 & 73.4 & 83.6 & 86.6 & 79.0 & 84.1 & 84.1 & \textbf{74.9} & 67.9 & 85.4 & 81.6 & 75.9 & 90.3 & 80.6 \\
0.5 & 73.5 & 83.6 & \textbf{87.2} & 78.9 & 84.6 & 85.0 & 74.4 & 67.8 & 85.4 & 81.4 & \textbf{76.4} & 90.4 & 80.7 \\
0.7 & \textbf{73.9} & \textbf{84.0} & 86.9 & \textbf{79.4} & \textbf{85.1} & \textbf{85.5} & 74.5 & \textbf{68.6} & \textbf{85.5} & \textbf{82.2} & 75.9 & \textbf{90.5} & \textbf{81.0}\\
\bottomrule
\end{tabular}
}
\end{table*}

\subsubsection{Performance of the EUDA Framework Under Various Configurations}
{\color{black}
In this experiment, we test our model in different configurations across various datasets. Our goal is to find which model best suits each dataset. Our findings indicate that while more complex models perform better on complex datasets, our simpler models, which have significantly fewer trainable parameters, can also achieve comparable results. This flexibility allows users to adjust the model's complexity to match their datasets' specific requirements. This adaptive capability is a distinctive feature of our approach which provides a unique advantage by offering a scalable solution that adjusts to varying data complexities without sacrificing performance. 

Table \ref{table_ours_oh} demonstrates our model’s performance on the Office-Home dataset with B and L feature extractors and S, B, L, and H bottleneck configurations. Our testing on Office-Home helped us identify optimal configurations, which reducing the need for extensive trials on other datasets. The L feature extractor was notably effective due to its ability to handle the significant domain variance within the dataset, and benefits from higher-dimensional features that capture more informative details. The LL configuration provides the best balance of complexity and performance.

Table \ref{table_ours_o31} shows the results of Office-31 dataset using B and L feature extractors and B and L bottleneck configurations. It can be seen that LL configuration produces the best results. Insights from the Office-Home tests informed the decision not to use the H bottleneck, as higher complexity had not resulted in improved performance in previous tests.

For the VisDA-2017 dataset (see Table \ref{table_ours_visda}), the BH configuration stood out, particularly suited to managing the transition from simulation to reality. This confirms the benefit of a more complex bottleneck in handling extensive domain shifts between real and simulation data. This emphasizes the importance of matching architectural choices to specific dataset challenges.

While we have conducted extensive testing across multiple configurations and datasets, time constraints limited the breadth of our experiments. We therefore encourage researchers and practitioners to further explore various bottleneck configurations to meet the specific demands and complexities of their datasets, enabling customized solutions that optimally address their unique challenges.
}

\section{Conclusion}
\label{sec:conclusion}
In this paper, we highlighted the potential of using a self-supervised pre-trained ViT for UDA by introducing a versatile framework that maintains simplicity, efficiency, and adaptability to ensure its applicability in practical scenarios. Specifically, we leveraged DINOv2, a self-supervised learning method in ViTs, to extract general features, and employed a simple yet effective bottleneck of fully connected layers to refine the extracted features. Additionally, we utilized the MMD loss to effectively align the source and target domains. Our model produces comparable results to state-of-the-art UDA methods with significantly fewer trainable parameters. This makes our method particularly suitable for real-world applications, including on-edge devices. Our proposed framework demonstrates promising results, achieving top-tier performance with 43\% to 99.7\% fewer trainable parameters across benchmark datasets compared to other methods. In our future research, we will explore additional UDA approaches based on self-supervised pre-trained ViT backbones and expand the applications of UDA in fields such as Autonomous Vehicles and other demanding areas where UDA is crucial.

\section*{Acknowledgements}
This research is partially funded by the NSERC CREATE TrustCAV program and the NSERC Discovery Grant program, and in part by the NSERC Canada Research Chair (CRC) Program.

\section*{Declarations}
\textbf{Conflict of interest.} The authors declare no competing interests.

\bibliography{references}

\end{document}